\pdfoutput=1

\documentclass[11pt]{article}

\usepackage{ACL2023}
\usepackage{booktabs}
\usepackage{times}
\usepackage{latexsym}
\usepackage{graphicx} 
\usepackage{subcaption}
\usepackage[T1]{fontenc}

\usepackage[utf8]{inputenc}

\usepackage{microtype}

\usepackage{inconsolata}

\usepackage{hyperref}

\title{Building Real-World Meeting Summarization Systems using \\ Large Language Models: A Practical Perspective
}

\author{Md Tahmid Rahman Laskar, Xue-Yong Fu, Cheng Chen, Shashi Bhushan TN \\
          Dialpad Canada Inc. \\
  \texttt{\{tahmid.rahman,xue-yong,cchen,sbhushan\}@dialpad.com}}

\begin{document}
\maketitle
\begin{abstract}
This paper studies how to effectively build meeting summarization systems for real-world usage using large language models (LLMs). For this purpose, we conduct an extensive evaluation and comparison of various closed-source and open-source LLMs, namely, GPT-4, GPT-3.5, PaLM-2, and LLaMA-2. Our findings reveal that most closed-source LLMs are generally better in terms of performance. However, much smaller open-source models like LLaMA-2 (7B and 13B) could still achieve performance comparable to the large closed-source models even in zero-shot scenarios. 
Considering the privacy concerns of closed-source models for only being accessible via API, alongside the high cost associated with using fine-tuned versions of the closed-source models, the open-source models that can achieve competitive performance are more advantageous for industrial use. Balancing performance with associated costs and privacy concerns, the LLaMA-2-7B model looks more promising for industrial usage. 
In sum, this paper offers practical insights on using LLMs for real-world business meeting summarization, shedding light on the trade-offs between performance and cost. 

\end{abstract}
\section{Introduction}
Meetings are a widely used method for collaboration, with 55 million meetings occurring each week in the USA \cite{zhong2021qmsum,hu-etal-2023-meetingbank}. With the rise of remote work, meetings have become even more crucial. While the widespread use of video conferencing software has made it easier to record meetings, the huge number of meetings makes it challenging to keep up with the information exchanged during them, emphasizing the need for automated methods of accessing key information. To address this issue, a summarization system that generates text summaries from meeting transcripts can be highly beneficial. However, the state-of-the-art neural summarization models \cite{bart, pegasus, t5} require large domain-specific summarization datasets for model training, which is difficult to obtain in real-world industrial scenarios due to the lack of domain-specific annotated data.


Recently, the impressive capability of LLMs to solve a wide range of tasks even in zero-shot scenarios \cite{laskar2023systematicchatgpt,qin2023chatgpt,bang2023multitask} has drawn a lot of interest among practitioners to apply LLMs to solve real-world problems. However, despite the effectiveness of LLMs across several tasks, there is a scarcity of comparative analysis between LLMs in long conversational data, especially in tasks such as meeting summarization. Thus, an extensive evaluation of LLMs in long meeting transcripts is an important step for the development of a real-world meeting summarization system that leverages LLM technologies. In this regard, this paper aims to provide a comprehensive analysis of various LLMs, which includes closed-source models like GPT-3.5 (i.e., ChatGPT\footnote{\url{https://openai.com/blog/chatgpt}}), GPT-4 \cite{openai2023gpt4},  and PaLM-2 \cite{palm2}, and two models (7B and 13B) of an open-source LLM: LLaMA-2 \cite{touvron2023llama2}. 
Our investigation focuses not only on the summarization quality of these models but also considers the cost-effectiveness and computational demands, providing a practical perspective to use such models in the real world.

Our experimental results show that while most closed-source models generally achieve better performance in meeting summarization datasets, the open-source LLaMA-2 models still achieve comparable performance while being significantly smaller. For this reason, an extensive model like GPT-4, despite its marginally superior performance, is deemed not as cost-effective due to its substantially higher computational requirements, alongside posing privacy risks since such closed-source models are only accessible via API. Furthermore, using fine-tuned versions of these closed-source models also significantly increases the API usage cost\footnote{\url{https://openai.com/blog/gpt-3-5-turbo-fine-tuning-and-api-updates}}. Thus, the trade-off between performance, cost, and computational demands makes open-source models like LLaMA-2 more favorable for industrial deployment. The insights from our research shed light on the practicalities of using LLMs for summarizing meeting transcripts. Thus, providing valuable guidance for similar industrial applications. Considering various factors, we employ LLaMA-2-7B in a real-world setting to generate summaries from Automatic Speech Recognition (ASR)-generated transcripts \cite{fu-etal-2022-entity,khasanova2022developing,laskar2022auto,laskar-etal-2022-blink,laskar-etal-2023-ai-coach-assist,manderscheid-lee-2023-predicting} of organizational meetings. Below, we summarize our major contributions in this paper:

(i) We conduct an extensive evaluation of closed-source LLMs as well as open-source LLMs in several benchmark meeting summarization datasets. 

(ii) We also present two approaches to address the long input sequence length issue in LLMs. 

(iii) Finally, we demonstrate a practical perspective on the trade-offs that come with selecting a model for real-world usage based on its performance, cost, and computational requirements.


\section{Related Work}
In recent years, neural models based on the transformer architecture \cite{DBLP:conf/nips/VaswaniSPUJGKP17} have led to state-of-the-art performance across various summarization datasets \cite{bart,pegasus,t5}. However, these transformer-based models require domain-specific fine-tuning \cite{devlin2019bert} to achieve the optimum performance. Thus, in real-world scenarios where in-domain labeled datasets are difficult to obtain, directly applying pre-trained transformers for zero-shot summarization may not yield great performance. In this regard, the impressive zero-shot performance of LLMs across various summarization datasets \cite{laskar2023systematicchatgpt} has drawn interest among practitioners to build real-world summarization systems via leveraging LLMs. 

In the real world, one interesting application of summarization is generating concise notes of long organizational meetings. Though several datasets \cite{janin2003icsi,carletta2005ami,gliwa2019samsum,clifton2020100spotify,chen2021dialogsum,khalman2021forumsum,zhu2021mediasum,cho2021streamhover,chen2022summscreen,nedoluzhko2022elitr} have been studied for the meeting summarization task in recent years, most of these datasets are not related to the business/organizational context. This makes these datasets less relevant for the evaluation of summarization models that require the generation of summaries from long organizational meetings. In this regard, some notable exceptions are the AMI \cite{carletta2005ami} dataset, the ICSI \cite{janin2003icsi} dataset, the MeetingBank \cite{hu-etal-2023-meetingbank} dataset, and the QMSUM \cite{zhong2021qmsum} dataset, as they consist of organizational meeting transcripts, contrary to other datasets. 

Since the development of a summarization system in a real-world industrial setting requires an extensive evaluation of the summarization model to ensure customer satisfaction, in this paper, we evaluate various LLMs in benchmark meeting summarization datasets. More specifically, we use the following datasets: AMI, ICSI, and QMSUM. All these datasets are constructed from organizational meetings. 
We hope that our extensive evaluation of LLMs in these meeting summarization datasets will help researchers and industry professionals to harness the power of LLMs to build real-world meeting summarization systems.

\section{Our Methodology}

\begin{figure*}
  \centering
  \includegraphics[width=\linewidth]{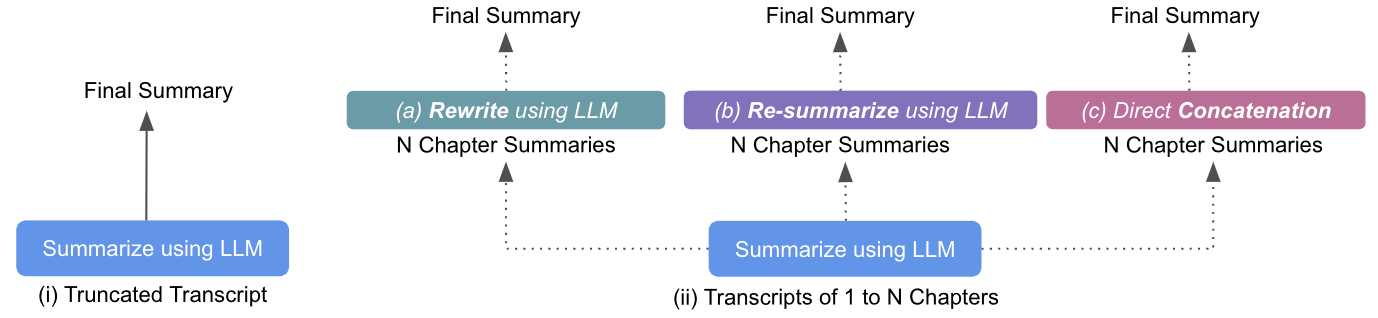} 
  \caption{\small{An overview of our proposed approaches. On the left (i), we demonstrate the \textit{Summarization via Truncation} approach where only the first $n$ words of the whole transcript are given as input to generate the summary. On the right (ii), we demonstrate the \textit{Summarization via Chapterization} approach where the summaries of every $n$ words of the transcript are first generated, denoted as chapter summaries, and then the final summary is generated either by (a) re-writing, (b) re-summarizing, or (c) concatenating the chapter summaries.}}
  \label{fig:overview} 
\end{figure*}

The objective of this research is to identify the most effective model to summarize organizational meetings that could be used in real-world applications in scenarios when in-domain labeled datasets are not available. Therefore, in this paper, we study LLMs due to their impressive zero-shot capabilities. However, one major issue with existing LLMs is their limitation to handle long contexts \cite{liu2023lost}. Since organizational meetings are usually quite longer \cite{zhong2021qmsum}, the input limit restrictions of LLMs make it difficult to consider the whole transcript sequence. In this regard, we study how to handle the long input sequence issue in LLMs based on the following two approaches (an overview of our proposed approaches is also shown in Figure \ref{fig:overview}):


\textbf{(i) Summarization via Truncation:} In this approach, we handle the input length restrictions in LLMs by only giving the first $n$ words of the meeting transcripts as input to the LLM for summarization. Our prompt for the LLM is given below:

{\textit{\textbf{Prompt:} ``Summarize the following conversation: [TRUNCATED TRANSCRIPT]''}}

\textbf{(ii) Summarization via Chapterization:} In our previous approach, the context of the whole transcript could not be given as input to the LLMs since only the truncated input sequence, consisting of the first $n$ words is provided. While this approach is quite efficient in terms of cost, the performance of the summarization may not be optimal due to truncating the input sequence length. To address this issue, we propose \textit{summarization via chapterization}. In this approach, we sequentially summarize every $n$ words. Here, we denote every $n$ words as chapters and the summary generated in each chapter as chapter summary. Afterward, we generate the final summary from these chapter summaries, either by concatenating the chapter summaries or by re-summarizing/re-writing the chapter summaries. To generate the summary of each chapter, we use the same prompt as we used in our previous approach of \textit{summarization via truncation}. To generate the final summary from the chapter summaries, in addition to directly \emph{\textbf{concatenating}} the chapter summaries, we investigate the following prompts: 

{\textit{\textbf{Prompt (Re-write):} ``Rewrite the following text by maintaining coherency: [Chapter Summaries]''}}

{\textit{\textbf{Prompt (Re-summarize):} ``Summarize the following text: [Chapter Summaries]''}}


\section{Experiments}
In this section, we first present our models along with their implementation details. Next, we demonstrate the datasets we used for evaluation. Finally, we demonstrate our experimental findings. 
\subsection{Models} 
We use four different LLMs (three closed-source and one open-source) to benchmark their performance in meeting transcripts. For the open-source model, we run our experiments in a \textit{g2-standard-96} machine in Google Cloud Platform\footnote{\url{https://cloud.google.com/}} (GCP), having 384GB of RAM memory and 8 NVIDIA L4 GPUs (24GB). For the closed-source models, we use their respective APIs.  Below, we describe these models. 

\textbf{GPT-3.5:} It is an autoregressive LLM that leverages the reinforcement learning from human feedback (RLHF) mechanism. 
It is the first backbone model behind ChatGPT and obtains impressive zero-shot performance across various tasks \cite{laskar2023systematicchatgpt}. 
We use the \textit{gpt-3.5-turbo-0613} model from OpenAI\footnote{\url{https://platform.openai.com/docs/models/}} that has the maximum context length of 4K tokens.  


\textbf{GPT-4:} It is the latest LLM released by OpenAI which is also the first multimodal model in the GPT series \cite{openai2023gpt4}. It is considered more reliable having better instruction-following capabilities than GPT-3.5. We use the version of \textit{gpt-4} that can consider the context length of 8k tokens. 


\textbf{PaLM-2:} PaLM-2 is an LLM \cite{palm2} developed by Google. It leverages the mixture of objectives technique \cite{palm2} and significantly outperforms the original PaLM \cite{chowdhery2022palm} model.  We use the \textit{text-bison@001 model in \textit{Google's VertexAI\footnote{\url{https://cloud.google.com/vertex-ai/docs/generative-ai/model-reference/text}}}} for PaLM-2 that has an input context window of 8K tokens.



\textbf{LLaMA-2:} LLaMA-2 \cite{touvron2023llama2} is an open-source LLM developed by Meta. One major advantage of LLaMA-2 over the previously mentioned LLMs is that it is open-sourced and available for both research and commercial purposes. In this paper, we use the respective Chat versions of LLaMA-2 for both 7B and 13B models from HuggingFace\footnote{\url{https://huggingface.co/meta-llama}} \cite{wolf2019huggingface}. 

\subsection{Datasets}

 \begin{table}[t!]
 \setlength{\tabcolsep}{4pt} 
\centering
\tiny
\begin{tabular}{lccc}
\toprule 
\textbf{Dataset} & \textbf{No. of Meetings} & \textbf{Avg. Transcript Len.} & \textbf{Avg. Summary Len.}   \\  \midrule
QMSUM\textsubscript{Filtered} & 37 & 10546 & 108 \\ \midrule
AMI & 20 & 7194 & 292 \\ \midrule
ICSI & 6 & 13998 & 454 \\ 
\bottomrule
\end{tabular}
\caption{Evaluation Dataset Statistics.}
\label{table:dataset_stat}
\end{table}


As our goal is to build the summarization system for real-world ASR-generated transcripts in the organizational meeting domain, we use datasets that are more relevant to our use-case. Note that for all datasets, we use their respective test sets and evaluate different LLMs using the same zero-shot prompts as our goal is to build a real-world meeting summarization system in scenarios when training datasets are not available. Thus, more generalized performance across various datasets is prioritized over obtaining state-of-the-art results on a particular dataset. Below, we describe these datasets (also see Table \ref{table:dataset_stat}).


\textbf{QMSUM dataset:} It is a meeting summarization dataset that consists of 232 meetings in multiple domains \cite{zhong2021qmsum}. However, this dataset is particularly constructed to conduct query-based summarization \cite{,laskar2022domain} of meetings, while our objective is to build a real-world summarization system that is required to generate the overall meeting summaries. Thus, we exclude instances from the QMSUM dataset that require the generation of meeting summaries depending on specific queries. 
This results in a filtered version of the QMSUM dataset (contains only 37 samples in the test set) that is more relevant to our target task. 


\textbf{AMI dataset:} The AMI \cite{carletta2005ami} dataset contains 140 scenario-based product design related meetings. The length of each meeting is usually between 30-40 minutes.

\textbf{ICSI dataset:} The ICSI \cite{janin2003icsi} dataset consists of 75
meetings. The length of each meeting in this dataset is approximately 1 hour. This dataset consists of research-related discussions among students at the International Computer Science Institute (ICSI) in Berkeley. 




\subsection{Results \& Discussions}
For performance evaluation, we use ROUGE-1, 2, L (R-1, R-2, R-L) \cite{lin2004rouge}, and BERTScore (B-S) \cite{zhang2019bertscore} as our evaluation metrics. For B-S, we use the DeBERTa-xlarge-mnli \cite{hedeberta} model. Below, we present our findings. 


\begin{table*}
\centering
\tiny
\begin{tabular}{l|l|cccc|cccc|cccc}
\toprule
\multicolumn{1}{c}{} & \multicolumn{1}{c}{} & \multicolumn{12}{c}{\textbf{Dataset}} \\ \cmidrule(r){3-14}  
\multicolumn{1}{c}{\textbf{Model}} & \multicolumn{1}{c}{\textbf{Type}} &  \multicolumn{4}{c}{\textbf{QMSUM (Filtered)}} &  \multicolumn{4}{c}{\textbf{AMI}} &  \multicolumn{4}{c}{\textbf{ICSI}} \\
\cmidrule(r){3-6} \cmidrule(r){7-10} \cmidrule(r){11-14} 

\multicolumn{1}{c}{} & \multicolumn{1}{c}{}
& \textbf{R-1} &  \textbf{R-2} &   \textbf{R-L} &   \textbf{B-S} & \textbf{R-1} &  \textbf{R-2} &   \textbf{R-L} &   \textbf{B-S} & \textbf{R-1} &  \textbf{R-2} &   \textbf{R-L} &   \textbf{B-S} \\
\midrule

GPT-3.5     & chapter (concat)
& 23.01    & 6.62  & 14.12  & 57.41  & 39.74 & {9.93} &           {19.71} &         {58.94} &                         36.01    &                           \textbf{7.56} &                15.21  &                 \textbf{57.32} \\
GPT-3.5      & chapter (resummarize) 
& 30.71  & 6.20  & 18.61  & {61.02 }   & 29.36 &  5.95 &           16.08 &          57.83 &                         23.43 &                           3.03 &                11.54 &                 53.46 \\
GPT-3.5     & chapter (rewrite)     
& 27.31  & 6.70  & 15.12  & 58.21  & {39.68} &  \textbf{9.94} &           \textbf{19.72} &          {59.71} &                         \textbf{37.50}  &                      {7.55}&                \textbf{16.29} &                 57.25 \\
GPT-3.5     & truncation       
& 32.01    & 6.62  & \textbf{19.02  }  & 60.81  & 29.63 &  6.52 &           16.43 &          57.85 &                         20.42 &                           2.71 &                10.95 &                 52.87 \\
\midrule
GPT-4       & chapter (concat) 
& 27.60 &	6.71 &	16.45 &	59.39  & 39.36	& 9.18	& 17.73 &	59.61   & 34.50	& 6.28	& 14.97&	57.05  \\

GPT-4        & chapter (resummarize) 
& 32.11  & 6.11  & 18.41  & \textbf{61.52}  & 30.02 & 6.56 &           15.96 &          57.88 &                         21.84 &                           3.89 &                11.86 &                 55.64 \\

GPT-4       & chapter (rewrite)     
& 30.05   & 7.06  & 17.07  & 60.13  & \textbf{39.76} &  9.65  &           19.25 &          \textbf{59.76} &                         36.39 &                           7.52 &                16.17 &                 57.28 \\
GPT-4       & truncation      
& 
\textbf{33.41} & \textbf{7.30}  & 17.82  & 60.91  & 32.56 &  6.75 &           16.93 &          58.01 &                         20.42 &                           4.23  &                12.02&                 53.64 \\
\midrule
 
   PaLM-2      & chapter (concat)
   & 20.61  & 4.12  & 11.92  & 48.92  & 16.11 &  1.01    &           11.35 &          47.08 &                         15.12 &                           1.24 &                11.27 &                 43.59 \\
 PaLM-2       & chapter (resummarize) 
 & 16.62  & 3.50  & 10.32  & 46.01    &  7.26 &  0.64 &            5.59 &          37.97 &                          5.31 &                           0.37 &                 3.75 &                 37.25 \\

PaLM-2      & chapter (rewrite)   
& 22.01 & 4.20  & 13.21  & 51.23  &  8.56 &  0.84 &            5.86 &          41.47 &                          8.58 &                           0.39 &                 5.83 &                 36.57 \\
PaLM-2      & truncation      
& 13.92  & 2.51  &  9.13  & 45.62  & 18.36 &  2.82 &           10.82 &          45.93 &                          8.43 &                           0.95 &                 6.07 &                 42.93 \\
\midrule

LLaMA-2-13b & chapter (concat) 
& 15.38 & 4.54 & 10.19 & 51.93 & 34.85 &  8.95 &           18.23 &          55.88 &                         32.31 &                           6.75 &                14.27 &                 53.97 \\

LLaMA-2-13b  & chapter (resummarize) 

& 29.01 & 5.71 & 17.64 & 55.49 & 28.73 &  6.28 &           16.61 &          54.71 &                         26.84 &                           4.42  &                13.31 &                 54.32 \\

 LLaMA-2-13b & chapter (rewrite)     
 & 26.73 & 6.33 & 16.83 & 54.37 & 37.38 &  8.37 &           19.36 &          57.55 &                         33.53 &                           6.05 &                15.06 &                 54.42 \\
LLaMA-2-13b & truncation      

& 28.64 & 6.39 & 18.29 & 55.32 & 33.38 &  7.24 &           18.64 &          55.38 &                         24.62 &                           3.35 &                13.39 &                 51.58 \\
\midrule

LLaMA-2-7b  & chapter (concat) 
& 15.72  & 4.37 & 10.03 & 51.93 & 32.34 &  8.08 &           16.33 &          53.91 &                         32.42 &                           7.21 &                13.82  &                 55.06 \\

 LLaMA-2-7b   & chapter (resummarize)
 & 29.65 & 6.37 & 17.41 & 57.66 & 30.92  &  5.95 &           16.63 &          56.63 &                         24.72  &                           4.45 &                12.17 &                 46.13 \\
 
 LLaMA-2-7b  & chapter (rewrite)    
 & 27.99 & 6.25 & 17.21 & 56.17 & 37.62  &  8.41 &           18.43  &          56.35 &                         26.59 &                           4.11 &                12.39 &                 48.52 \\

 LLaMA-2-7b  & truncation       
 
 & 25.48 & 5.69 & 15.01 & 53.58 & 30.22  &  6.59 &           16.52 &          55.35 &                         17.72 &                           2.14 &                 9.24 &                 48.84 \\

\bottomrule

\end{tabular}
\caption{\small{Performance of LLMs on the QMSUM (Filtered), AMI, and ICSI Datasets.}}
\label{tab:results}
\end{table*}

\subsubsection{Performance on Benchmark Datasets}
\begin{figure*}[t!]
    \centering
    
    \begin{subfigure}[t!]{\textwidth} 
        \includegraphics[width=\textwidth]{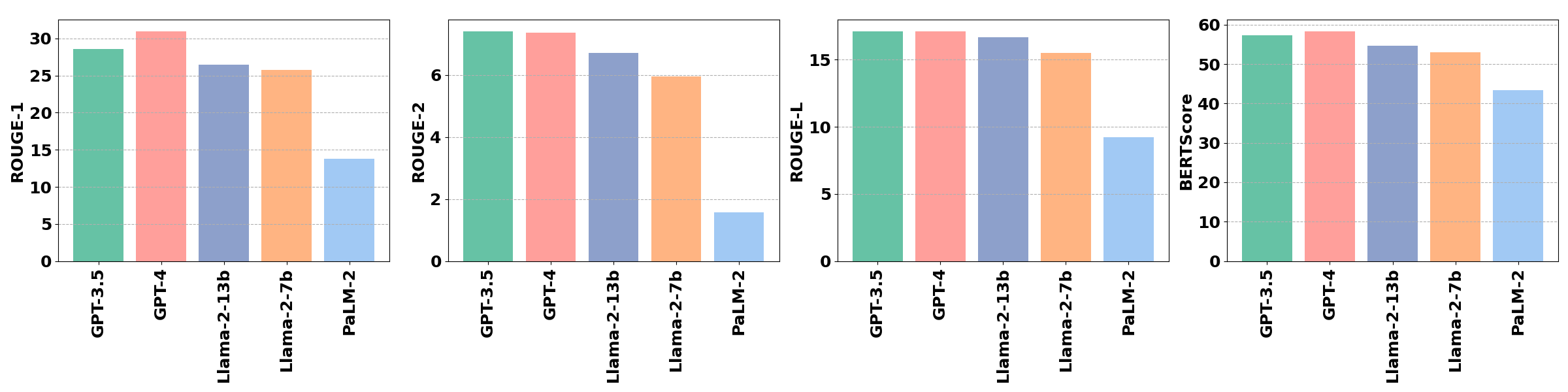}
        \caption{Based on different model types.}
        \label{fig:average_results_models}
    \end{subfigure}
    
    \begin{subfigure}[t!]{\textwidth}
        \includegraphics[width=1\textwidth]{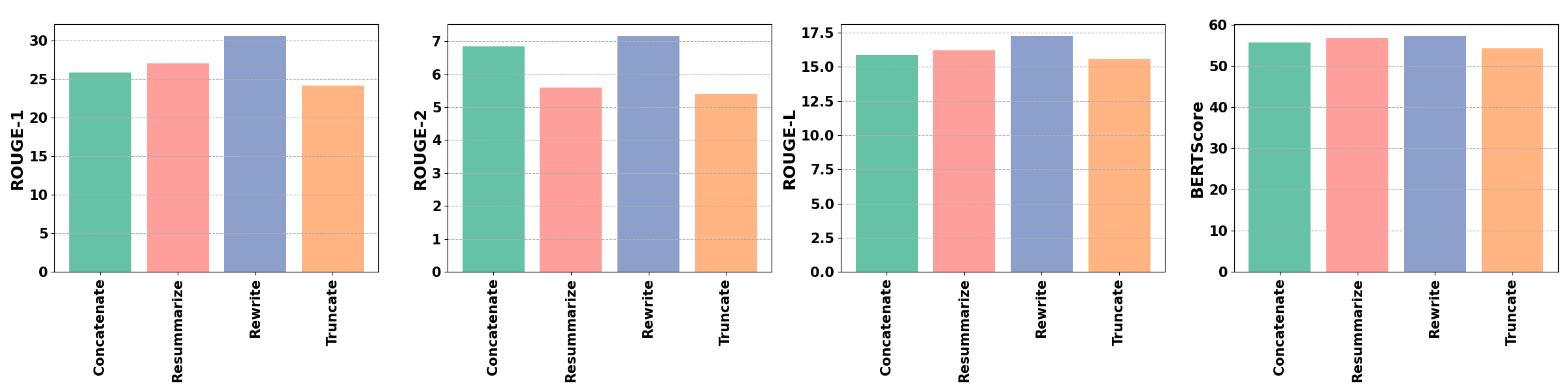}
        \caption{Based on different summarization types.}
        \label{fig:average_results_types}
    \end{subfigure}
    
    \caption{Average score across all datasets based on (a) \textit{model types} and (b) \textit{summarization types}.}
    \label{fig:avg}
\end{figure*}

While most of the LLMs we use in this paper for evaluation support at least 4K tokens, we find that longer sequence length makes the inference slower, which is not practical as our goal is to build a summarization system for production usage. Also, in terms of open-source models like LLaMA-2, it increases the computational requirements (e.g., requires high-end GPUs). Considering these restrictions, we use $n=2500$ words as the maximum input sequence length for all models to ensure a fair evaluation. 
We show the results for all LLMs in different datasets in Table \ref{tab:results} (For simplicity, we also demonstrate the performance based on the average across all datasets according to various summarization types and different models in Figure \ref{fig:avg}). 
Below, we summarize our observations:

(i) In the QMSUM (Filtered) dataset, we surprisingly find that the  \textit{summarization via chapterization} approaches fail to outperform the \textit{summarization via truncation} approach in many scenarios. This is surprising since the model does not have access to the whole context when the \textit{summarization via truncation} approach is used, indicating that the gold reference summaries in this dataset could possibly be more biased towards the beginning of the transcript. 

  (ii) In the AMI and ICSI datasets, among the closed-source models, the performance of the PaLM-2 model noticeably lags behind that of GPT-3.5, and GPT-4. More interestingly, PaLM-2 even lags behind much smaller LLaMA-2-7B and LLaMA-2-13B models. Note that we did not observe such a poor performance by PaLM-2 in the QMSUM dataset. 
  
  (iii) In the AMI dataset, we find that GPT-4 is the best in terms of ROUGE-1 and BERTScore, while GPT-3.5 is found to be the best in terms of ROUGE-2\&L. For both models, the best results in these metrics are achieved based on the \textit{chapterization via re-writing} approach. However, we find that the performance is generally much poorer for these models when the \textit{chapterization via re-summarization} and the \textit{summarization via truncation} approaches are used (we also find quite similar trends for the LLaMA-2 models in this dataset). 
    
    (iv) In the ICSI dataset, we find that various approaches using GPT-3.5 led to the best performance across different metrics. In terms of ROUGE-1\&L, we find that \textit{chapterization via re-writing} is the best, while in terms of ROUGE-2 and BERTScore, \textit{chapterization via concatenation} led to the best result. Similar to the AMI dataset, we again observe poor performance for GPT and LLaMA models using the \textit{chapterization via re-summarization} and  \textit{summarization via truncation} approaches in the ICSI dataset, with truncation-based approach leading to the poorest performance. This may indicate that in AMI and ICSI datasets that require longer summaries (approximately, 300-450 words long summaries on average), either concatenation or re-writing of the chapter summaries is more useful. 

    
     (v) In the QMSUM dataset, the \textit{truncation} approach is found to be the best-performing approach, as GPT-4 using this technique achieves the best result in terms of ROUGE-1\&2, while GPT-3.5 achieves the best based on ROUGE-L. However, in terms of BERTScore, we find that GPT-4 based on \textit{chapterization via resummarization} approach to be the best. These findings may indicate that in datasets where the average gold reference summary length is smaller (in QMSUM, the average summary length is just 108 words), \textit{chapterization via resummarization} or \textit{summarization via truncation} approaches are more useful.

    (vi) The average across all datasets in terms of different LLMs (see Figure \ref{fig:average_results_models}) demonstrate that GPT-4 is generally the best, followed by GPT-3.5, with PaLM-2 being the worst performer. We also observe that LLaMA-2 models achieve competitive performance, with the 7B model performing almost similar to the 13B model. This opens up the possibility of fine-tuning smaller LLMs on in-domain datasets to achieve further performance gain. 

    (vii) Based on the average across all datasets in terms of different summarization types (see Figure \ref{fig:average_results_types}), the \textit{chapterization via rewriting} approach is found to be the best. We also do not find any significant difference in performance based on various summarization types. Since the truncation-based approach achieves performance comparable to various chapterization-based approaches even without considering the whole context, further investigation 
    of the gold summaries in these datasets is needed.

 \begin{table}[t!]
\centering
\tiny
\setlength{\tabcolsep}{4pt}
\begin{tabular}{c|cccc|cccc}
\toprule 
\multicolumn{1}{c}{}& \multicolumn{8}{c}{\textbf{Context Length}}  \\ 
\cmidrule{2-9}
    \multicolumn{1}{c}{} & \multicolumn{4}{c}{\textbf{2500}} & \multicolumn{4}{c}{\textbf{5000}}  \\ 
     \cmidrule(r){2-5} \cmidrule(r){6-9}  
   \textbf{Dataset}  & \textbf{R-1} &  \textbf{R-2} &   \textbf{R-L} &   \textbf{B-S} & \textbf{R-1} &  \textbf{R-2} &   \textbf{R-L} &   \textbf{B-S}  \\  \midrule

{QMSUM} & 33.41 & 7.30 & 17.82 & 60.91 & 33.43 & 7.69 & 18.30 & 61.02 \\
{AMI} & 32.56 & 6.75 & 16.93 & 58.01 & 30.59 & 5.68 & 14.94 & 56.97 \\
{ICSI} & 20.42 & 4.23 & 10.92 & 53.64 & 24.32 & 4.41 & 12.04 & 54.53 \\
\bottomrule
\end{tabular}
\caption{Results based on Context Length for GPT-4.} 
\label{table:context_variation_performance}
\end{table}

%

\subsubsection{Case Study}

To further investigate the performance of LLMs, we conduct some case studies using the \textit{Summarization via Truncation} approach, as demonstrated below.

\paragraph{(i) Case Study on Sequence Length:} In this case study, we investigate how the value of $n$ for maximum input sequence length impacts the overall performance. In our study, we use the GPT-4 model and increase the value of $n$ from 2500 to 5000. We present the results in different datasets in Table \ref{table:context_variation_performance} to find that increasing the maximum input sequence length from 2500 to 5000 does not necessarily lead to an increase in performance. 
More specifically, in datasets that have an average transcript length of more than 10,000 words, we observe that the performance is increased with the increase in context length (e.g., QMSUM\textsubscript{Filtered} and ICSI). While the performance drops in datasets having smaller context length (e.g., AMI). \citet{liu2023lost} also find that increasing the sequence length of LLMs to consider long contexts does not necessarily improve the performance. Thus, future work should conduct more qualitative evaluations to identify the reason behind this trend. 

\paragraph{(ii) Case Study on Prompt Variations:} Here, we conduct experiments with GPT-3.5 in the QMSUM dataset using the following prompts:


\textbf{Long:} Generate a long and descriptive summary of the following conversation.

\textbf{Medium:} Generate a summary of the following conversation.

\textbf{Short:} Generate a very short and concise summary of the following conversation.

We present our results for this case study in Table \ref{table:prompt_variation_performance} to find that different prompts yield different results. For instance, prompting to generate long summaries led to an average summary length of 402.70, which also led to the poorest performance in terms of both ROUGE and BERTScore. 
Meanwhile, shorter summaries yield better performance. Since the average length of the summary in QMSUM is 108 words, the prompt to generate the summary without explicitly specifying the size (long or short) leads to an average summary length of 136 words, which also achieves the best performance. These findings demonstrate that based on user requirements, we may build summarization systems in the real-world that can generate summaries of various lengths. 

 \begin{table}[t!]
\centering
\tiny
\begin{tabular}{llllll}
\toprule 
    \textbf{Type} & \textbf{R-1} &  \textbf{R-2} &   \textbf{R-L} &   \textbf{B-S}  & \textbf{Length}  \\  \midrule

            Truncation (Long) & 23.61 &  5.68 & 13.64 & 55.99 &        402.70\\
          Truncation (Medium) & 32.01    & 6.62  & {19.02}  & 60.81  &        136.35 \\
          Truncation (Short) & 31.81 & 6.19 & 18.76 & 60.34 &        74.40\\

\bottomrule
\end{tabular}
\caption{Results based on Prompt Variations in the QMSUM (filtered) Dataset for GPT-3.5. Here, `Length' denotes ``average length of the generated summary''.} 
\label{table:prompt_variation_performance}
\end{table}



\section{Using LLMs in Real-World Systems}

To \textit{deploy} LLMs in the real world, we study the following aspects: \textit{cost} and  \textit{inference speed}.

\textbf{Cost:} Based on our experiments, we find that except for the PaLM-2 model, the closed-source LLMs usually outperform the open-source LLaMA-2 models. While GPT-4 generally performs the best, it is also more costly. As of the writing of this paper, the pricing\footnote{\url{https://openai.com/pricing}} in OpenAI for the GPT series models are as follows: for GPT-4, the 8K context version costs $0.03$\$ per 1K input tokens and $0.06$\$ per 1K output tokens, while for the 4K context version of GPT-3.5 that we use costs $0.0015$\$ per 1K input tokens and $0.002$\$ per 1K output tokens. On average, it makes GPT-4 25 times more costly than GPT-3.5. Meanwhile, for PaLM-2, the pricing\footnote{\url{https://cloud.google.com/vertex-ai/pricing}} in Google Cloud is $0.0010$\$ per 1000 characters (for both input and output). Approximately, 1 token is considered as 4 characters. Thus, the cost for PaLM-2 is $0.0040$\$ per 1K tokens, making it 
about 2.5 times more costly than GPT-3.5. 
Regarding LLaMA-2, we 
were able to run it in a machine with 1 NVIDIA L4 GPU, while the LLaMA-2-13B model was possible to run using 2 L4 GPUs. However, using multiple GPUs significantly increases the production cost.  Thus, in terms of real-world usage, LLaMA-2-7B is more useful than LLaMA-2-13B as we observe almost similar performance using these models across various datasets. 

\textbf{Inference Speed:} We also measure the inference speed of different LLMs. For this purpose, we collected 100 meeting transcripts from Dialpad\footnote{\url{https://www.dialpad.com/ca/}} that consist of real-world conversations. All the collected transcripts have at least 2500 words. Based on our analysis using the \textit{Summarization via Truncation} approach, we find that GPT-3.5 is the fastest, as it takes 2.5 seconds on average per transcript for inference, followed by PaLM-2 which takes about 3.2 seconds on average. While GPT-4 achieves the best performance in terms of ROUGE and BERTScore metrics, it is also the slowest among the commercial closed-source LLMs since it takes about 11 seconds on average per transcript. We also benchmark the inference speed of LLaMA-2-7B in GCP on a machine having 1 NVIDIA L4 GPU and find that it takes 15 seconds on average. 



\textbf{Deployment:} While using the APIs of closed-source models usually leads to faster inference speed, there are some drawbacks of using closed-source LLMs. For instance, businesses need to send their customer data using the 3rd-party API, which may lead to potential privacy risks. Meanwhile, we also observe that these LLM APIs sometimes become too slow when the demand is high, leading to API request failure. By considering such cases of API failures, the average inference speed of GPT-4 is increased to 40 seconds per transcript (up from 11 seconds). Thus, based on cost and inference speed trade-off, alongside the privacy concerns and the possibility of fine-tuning on in-domain datasets without additional deployment costs, LLaMA-2-7B looks more promising for production usage. Meanwhile, we can also leverage various model optimization  \cite{zhu2023survey} techniques like quantization, pruning, distillation, etc. to further reduce the memory requirement as well as improve the inference speed. Therefore, we select the LLaMA-2-7B model for real-world usage and after further fine-tuning on our in-domain data, we successfully deploy it in our production environment in a machine having 1 NVIDIA L4 GPU. 

\section{Conclusion} 
In this paper, our extensive study involving various LLMs led to several key insights on building a real-world meeting summarization system. While most closed-source LLMs usually outperform their open-source counterparts, striking a balance between cost and performance (while also addressing potential privacy risks) is crucial for practical, real-world deployment. This became evident in the case of GPT-4, which, while showing superior performance in most datasets, was considerably less cost-effective. 
By considering the performance and cost trade-off, as well as the privacy concern, we deploy a summarization model based on LLaMA-2-7B to generate live summaries of ASR-generated meeting transcripts.  
This research thereby provides crucial insights and a clear perspective on deploying LLMs in real-world industrial settings. 
In the future, we will investigate the performance of LLMs by applying various optimization techniques. 


\section*{Limitations}
One of the limitations of this work is that the models were only evaluated on academic datasets even though our focus was to build this system for real-world usage for a particular business organization to summarize meeting conversations. Thus, future work should focus on constructing a dataset for the target domain consisting of real-world conversations. Moreover, it has been found recently that existing evaluation metrics like ROUGE have several limitations while evaluating the performance of LLMs in summarization datasets \cite{laskar2023systematicchatgpt,goyal2022news}. Thus, future work should also benchmark the performance of LLMs in meeting summarization based on extensive human evaluation. While this paper evaluates 3 closed-source LLMs and 1 open-source LLM, there are many other recently proposed open-source LLMs \cite{zhang2023instruction,zhao2023survey,chang2023survey}, such as Cerebras\footnote{\url{https://huggingface.co/cerebras/Cerebras-GPT-13B}}, Pythia\footnote{\url{https://github.com/EleutherAI/pythia}}, Dolly\footnote{\url{https://huggingface.co/databricks/dolly-v2-7b}}, Vicuna\footnote{\url{https://huggingface.co/lmsys/vicuna-7b-v1.5}}, MPT\footnote{\url{https://huggingface.co/mosaicml/mpt-7b-instruct}}, RedPajama\footnote{\url{https://huggingface.co/togethercomputer/RedPajama-INCITE-7B-Instruct}}, Falcon\footnote{\url{https://huggingface.co/tiiuae/falcon-7b}}, XGen\footnote{\url{https://github.com/salesforce/xgen}},
Mistral\footnote{\url{https://huggingface.co/mistralai/Mistral-7B-Instruct-v0.1}}, which are not evaluated in this work. Nonetheless, LLaMA-2 is found to be one of the best-performing open-source models \cite{zhao2023survey} and so we select this model in our work. Though the performance of different LLMs may not be state-of-the-art in various datasets, the intended development of the summarization system using LLMs was to ensure more generalized performance in our targeted domain across various types of meetings, instead of obtaining state-of-the-art performance in a particular dataset.


\section*{Ethics Statement}

\paragraph{License:} We maintained the licensing requirements accordingly while using different tools from the providers (e.g., OpenAI, Google, Meta, HuggingFace). 

\paragraph{Privacy:} To protect user privacy, sensitive
data such as personally identifiable information (e.g., credit card number, phone number, person names)
were removed while benchmarking the inference speed of different LLMs on the collected 100 real-world transcripts. 

\paragraph{Intended Use:}  Note that our model is intended to provide business organizations with a quick overview of the meetings. While poor summarization quality may lead to a bad user experience, it should not lead to any ethical concern since the summary is required to be generated based on only the given transcript. Meanwhile, the LLM that would be used in production for summarization will only do inference but will not be re-trained on live meeting transcripts. Only the users of a particular meeting will have access to the summary and so information from any other meetings will not be revealed to the users. 

\appendix

\label{sec:appendix} 
\section{Appendix}

\subsection{Performance of LLMs on other Meeting Summarization datasets}

In this paper, while we primarily use the QMSUM, AMI, and ICSI datasets that are more relevant to our targeted domain for meeting summarization, to help future research on using LLMs in meeting summarization, we also benchmark the performance of different LLMs in another meeting summarization dataset: MeetingBank \cite{hu-etal-2023-meetingbank}. It is a recently proposed meeting summarization dataset that consists of city council meeting transcripts in the USA over the past decade. On average, a meeting segment in this dataset has 2892 tokens in the transcript and 87 tokens in the summary. While this dataset is not directly related to our targeted domain, we only conducted this experiment for research purposes to benchmark the performance of LLMs in related datasets to help further research on using LLMs for meeting summarization. We use all 862 instances of the test set for this experiment. We find from our experimental results presented in Table \ref{tab:results_meetingbank} that the \textit{chapterization via resummarization} approach using GPT-4 is the best-performing one in terms of ROUGE-1\&L, and BERTScore, while LLaMA-2-7B based on \textit{Truncation} is the best-performing approach in terms of ROUGE-2. A possible explanation behind this finding could be because the average length of the summaries in the MeetingBank test set is just 57 words, and so the \textit{Truncation} or the \textit{chapterization based on resummarization} approaches may end up generating smaller summaries. 

\begin{table}
\centering
\tiny
\begin{tabular}{l|l|cccc}
\toprule

\multicolumn{1}{c}{\textbf{Model}} & \multicolumn{1}{c}{\textbf{Type}}
& \textbf{R-1} &  \textbf{R-2} &   \textbf{R-L} &   \textbf{B-S}  \\
\midrule

GPT-3.5     & chapter (concat)
 & 22.78 &  9.06 & 16.42 & 54.76 
 \\
GPT-3.5      & chapter (resummarize) 
& 24.86 &  9.33 & 17.52  & 55.95 
 \\
GPT-3.5     & chapter (rewrite)     
 & 24.71 &  9.26 & 17.31 & 55.93 \\
GPT-3.5     & truncation       
& 25.63 &  9.90  & 18.31  & 56.24 
 \\
\midrule
GPT-4       & chapter (concat) 
& 25.76	& 10.61	& 18.29	& 57.58   \\

GPT-4        & chapter (resummarize) 
 & \textbf{27.02}    & 10.85 & \textbf{18.93} & \textbf{57.78} 
 \\

GPT-4       & chapter (rewrite)     
& 26.01 & 10.68 & 18.54 & 57.54 
\\
GPT-4       & truncation      
 & 26.82  & 10.87 & 18.68 & 57.75 
 \\
\midrule
 
   PaLM-2      & chapter (concat)
   & 22.01    &  9.34 & 16.68 & 53.65 
 \\
 PaLM-2       & chapter (resummarize) 
& 23.02 &  9.81 & 17.52 & 53.95 
 \\

PaLM-2      & chapter (rewrite)   
& 22.51 &  9.56 & 17.12 & 53.87 
\\
PaLM-2      & truncation      
 & 20.86 &  9.67 & 16.79 & 52.08 
\\
\midrule

LLaMA-2-13b & chapter (concat) 
& 13.44 &  7.76 & 10.77 & 51.52 
 \\

LLaMA-2-13b  & chapter (resummarize) 
& 26.61 & 10.53 & 18.86 & 56.82 
 \\

 LLaMA-2-13b & chapter (rewrite)     
 & 18.66 &  9.63 & 14.29  & 52.84 
\\
LLaMA-2-13b & truncation      
& 24.55 & 11.03 & 17.82 & 56.05 
 \\
\midrule

LLaMA-2-7b  & chapter (concat) 
& 13.49 &  7.66 & 10.73 & 51.59 
 \\

 LLaMA-2-7b   & chapter (resummarize)
 & 26.38 & 11.01 & 18.77 & 57.83 
 \\
 
 LLaMA-2-7b  & chapter (rewrite)    
& 20.17 &  9.41  & 14.92 & 54.23 \\

 LLaMA-2-7b  & truncation       
 & 24.75 & \textbf{11.08} & 17.95 & 56.49 \\

\bottomrule

\end{tabular}
\caption{\small{Performance of LLMs on the MeetingBank dataset.}}
\label{tab:results_meetingbank}
\end{table}

\bibliography{anthology,custom}

\begin{thebibliography}{42}
\expandafter\ifx\csname natexlab\endcsname\relax\def\natexlab#1{#1}\fi

\bibitem[{Bang et~al.(2023)Bang, Cahyawijaya, Lee, Dai, Su, Wilie, Lovenia, Ji, Yu, Chung, Do, Xu, and Fung}]{bang2023multitask}
Yejin Bang, Samuel Cahyawijaya, Nayeon Lee, Wenliang Dai, Dan Su, Bryan Wilie, Holy Lovenia, Ziwei Ji, Tiezheng Yu, Willy Chung, Quyet~V. Do, Yan Xu, and Pascale Fung. 2023.
\newblock \href {http://arxiv.org/abs/2302.04023} {A multitask, multilingual, multimodal evaluation of chatgpt on reasoning, hallucination, and interactivity}.

\bibitem[{Brown et~al.(2020)Brown, Mann, Ryder, Subbiah, Kaplan, Dhariwal, Neelakantan, Shyam, Sastry, Askell et~al.}]{gpt3}
Tom Brown, Benjamin Mann, Nick Ryder, Melanie Subbiah, Jared~D Kaplan, Prafulla Dhariwal, Arvind Neelakantan, Pranav Shyam, Girish Sastry, Amanda Askell, et~al. 2020.
\newblock Language models are few-shot learners.
\newblock \emph{Advances in neural information processing systems}, 33:1877--1901.

\bibitem[{Carletta et~al.(2005)Carletta, Ashby, Bourban, Flynn, Guillemot, Hain, Kadlec, Karaiskos, Kraaij, Kronenthal et~al.}]{carletta2005ami}
Jean Carletta, Simone Ashby, Sebastien Bourban, Mike Flynn, Mael Guillemot, Thomas Hain, Jaroslav Kadlec, Vasilis Karaiskos, Wessel Kraaij, Melissa Kronenthal, et~al. 2005.
\newblock The ami meeting corpus: A pre-announcement.
\newblock In \emph{International workshop on machine learning for multimodal interaction}, pages 28--39. Springer.

\bibitem[{Chang et~al.(2023)Chang, Wang, Wang, Wu, Zhu, Chen, Yang, Yi, Wang, Wang et~al.}]{chang2023survey}
Yupeng Chang, Xu~Wang, Jindong Wang, Yuan Wu, Kaijie Zhu, Hao Chen, Linyi Yang, Xiaoyuan Yi, Cunxiang Wang, Yidong Wang, et~al. 2023.
\newblock A survey on evaluation of large language models.
\newblock \emph{arXiv preprint arXiv:2307.03109}.

\bibitem[{Chen et~al.(2022)Chen, Chu, Wiseman, and Gimpel}]{chen2022summscreen}
Mingda Chen, Zewei Chu, Sam Wiseman, and Kevin Gimpel. 2022.
\newblock Summscreen: A dataset for abstractive screenplay summarization.
\newblock In \emph{Proceedings of the 60th Annual Meeting of the Association for Computational Linguistics (Volume 1: Long Papers)}, pages 8602--8615.

\bibitem[{Chen et~al.(2021)Chen, Liu, Chen, and Zhang}]{chen2021dialogsum}
Yulong Chen, Yang Liu, Liang Chen, and Yue Zhang. 2021.
\newblock \href {https://doi.org/10.18653/v1/2021.findings-acl.449} {{D}ialog{S}um: {A} real-life scenario dialogue summarization dataset}.
\newblock In \emph{Findings of the Association for Computational Linguistics: ACL-IJCNLP 2021}, pages 5062--5074, Online. Association for Computational Linguistics.

\bibitem[{Cho et~al.(2021)Cho, Dernoncourt, Ganter, Bui, Lipka, Chang, Jin, Brandt, Foroosh, and Liu}]{cho2021streamhover}
Sangwoo Cho, Franck Dernoncourt, Tim Ganter, Trung Bui, Nedim Lipka, Walter Chang, Hailin Jin, Jonathan Brandt, Hassan Foroosh, and Fei Liu. 2021.
\newblock \href {https://doi.org/10.18653/v1/2021.emnlp-main.520} {{S}tream{H}over: Livestream transcript summarization and annotation}.
\newblock In \emph{Proceedings of the 2021 Conference on Empirical Methods in Natural Language Processing}, pages 6457--6474, Online and Punta Cana, Dominican Republic. Association for Computational Linguistics.

\bibitem[{Chowdhery et~al.(2022)Chowdhery, Narang, Devlin, Bosma, Mishra, Roberts, Barham, Chung, Sutton, Gehrmann et~al.}]{chowdhery2022palm}
Aakanksha Chowdhery, Sharan Narang, Jacob Devlin, Maarten Bosma, Gaurav Mishra, Adam Roberts, Paul Barham, Hyung~Won Chung, Charles Sutton, Sebastian Gehrmann, et~al. 2022.
\newblock Palm: Scaling language modeling with pathways.
\newblock \emph{arXiv preprint arXiv:2204.02311}.

\bibitem[{Clifton et~al.(2020)Clifton, Reddy, Yu, Pappu, Rezapour, Bonab, Eskevich, Jones, Karlgren, Carterette et~al.}]{clifton2020100spotify}
Ann Clifton, Sravana Reddy, Yongze Yu, Aasish Pappu, Rezvaneh Rezapour, Hamed Bonab, Maria Eskevich, Gareth Jones, Jussi Karlgren, Ben Carterette, et~al. 2020.
\newblock 100,000 podcasts: A spoken english document corpus.
\newblock In \emph{Proceedings of the 28th International Conference on Computational Linguistics}, pages 5903--5917.

\bibitem[{Devlin et~al.(2019)Devlin, Chang, Lee, and Toutanova}]{devlin2019bert}
Jacob Devlin, Ming-Wei Chang, Kenton Lee, and Kristina Toutanova. 2019.
\newblock Bert: Pre-training of deep bidirectional transformers for language understanding.
\newblock In \emph{Proceedings of the 2019 Conference of the North American Chapter of the Association for Computational Linguistics: Human Language Technologies, Volume 1 (Long and Short Papers)}, pages 4171--4186.

\bibitem[{Fu et~al.(2022)Fu, Chen, Laskar, Gardiner, Hiranandani, and Tn}]{fu-etal-2022-entity}
Xue-yong Fu, Cheng Chen, Md~Tahmid~Rahman Laskar, Shayna Gardiner, Pooja Hiranandani, and Shashi~Bhushan Tn. 2022.
\newblock \href {https://doi.org/10.18653/v1/2022.emnlp-industry.49} {Entity-level sentiment analysis in contact center telephone conversations}.
\newblock In \emph{Proceedings of the 2022 Conference on Empirical Methods in Natural Language Processing: Industry Track}, pages 484--491, Abu Dhabi, UAE. Association for Computational Linguistics.

\bibitem[{Gliwa et~al.(2019)Gliwa, Mochol, Biesek, and Wawer}]{gliwa2019samsum}
Bogdan Gliwa, Iwona Mochol, Maciej Biesek, and Aleksander Wawer. 2019.
\newblock \href {https://doi.org/10.18653/v1/D19-5409} {{SAMS}um corpus: A human-annotated dialogue dataset for abstractive summarization}.
\newblock In \emph{Proceedings of the 2nd Workshop on New Frontiers in Summarization}, pages 70--79, Hong Kong, China. Association for Computational Linguistics.

\bibitem[{Google(2023)}]{palm2}
Google. 2023.
\newblock \href {https://ai.google/static/documents/palm2techreport.pdf} {Palm 2 technical report}.
\newblock \emph{Goole AI}.

\bibitem[{Goyal et~al.(2022)Goyal, Li, and Durrett}]{goyal2022news}
Tanya Goyal, Junyi~Jessy Li, and Greg Durrett. 2022.
\newblock News summarization and evaluation in the era of gpt-3.
\newblock \emph{arXiv preprint arXiv:2209.12356}.

\bibitem[{He et~al.()He, Liu, Gao, and Chen}]{hedeberta}
Pengcheng He, Xiaodong Liu, Jianfeng Gao, and Weizhu Chen.
\newblock Deberta: Decoding-enhanced bert with disentangled attention.
\newblock In \emph{International Conference on Learning Representations}.

\bibitem[{Hu et~al.(2023)Hu, Ganter, Deilamsalehy, Dernoncourt, Foroosh, and Liu}]{hu-etal-2023-meetingbank}
Yebowen Hu, Timothy Ganter, Hanieh Deilamsalehy, Franck Dernoncourt, Hassan Foroosh, and Fei Liu. 2023.
\newblock \href {https://aclanthology.org/2023.acl-long.906} {{M}eeting{B}ank: A benchmark dataset for meeting summarization}.
\newblock In \emph{Proceedings of the 61st Annual Meeting of the Association for Computational Linguistics (Volume 1: Long Papers)}, pages 16409--16423, Toronto, Canada. Association for Computational Linguistics.

\bibitem[{Janin et~al.(2003)Janin, Baron, Edwards, Ellis, Gelbart, Morgan, Peskin, Pfau, Shriberg, Stolcke et~al.}]{janin2003icsi}
Adam Janin, Don Baron, Jane Edwards, Dan Ellis, David Gelbart, Nelson Morgan, Barbara Peskin, Thilo Pfau, Elizabeth Shriberg, Andreas Stolcke, et~al. 2003.
\newblock The icsi meeting corpus.
\newblock In \emph{2003 IEEE International Conference on Acoustics, Speech, and Signal Processing, 2003. Proceedings.(ICASSP'03).}, volume~1, pages I--I. IEEE.

\bibitem[{Khalman et~al.(2021)Khalman, Zhao, and Saleh}]{khalman2021forumsum}
Misha Khalman, Yao Zhao, and Mohammad Saleh. 2021.
\newblock Forumsum: A multi-speaker conversation summarization dataset.
\newblock In \emph{Findings of the Association for Computational Linguistics: EMNLP 2021}, pages 4592--4599.

\bibitem[{Khasanova et~al.(2022)Khasanova, Hiranandani, Gardiner, Chen, Corston-Oliver, and Fu}]{khasanova2022developing}
Elena Khasanova, Pooja Hiranandani, Shayna Gardiner, Cheng Chen, Simon Corston-Oliver, and Xue-Yong Fu. 2022.
\newblock \href {https://doi.org/10.18653/v1/2022.naacl-industry.29} {Developing a production system for {P}urpose of {C}all detection in business phone conversations}.
\newblock In \emph{Proceedings of the 2022 Conference of the North American Chapter of the Association for Computational Linguistics: Human Language Technologies: Industry Track}, pages 259--267, Hybrid: Seattle, Washington + Online. Association for Computational Linguistics.

\bibitem[{Laskar et~al.(2023{\natexlab{a}})Laskar, Bari, Rahman, Bhuiyan, Joty, and Huang}]{laskar2023systematicchatgpt}
Md~Tahmid~Rahman Laskar, M~Saiful Bari, Mizanur Rahman, Md~Amran~Hossen Bhuiyan, Shafiq Joty, and Jimmy Huang. 2023{\natexlab{a}}.
\newblock \href {https://aclanthology.org/2023.findings-acl.29} {A systematic study and comprehensive evaluation of {C}hat{GPT} on benchmark datasets}.
\newblock In \emph{Findings of the Association for Computational Linguistics: ACL 2023}, pages 431--469, Toronto, Canada. Association for Computational Linguistics.

\bibitem[{Laskar et~al.(2023{\natexlab{b}})Laskar, Chen, Fu, Azizi, Bhushan, and Corston-oliver}]{laskar-etal-2023-ai-coach-assist}
Md~Tahmid~Rahman Laskar, Cheng Chen, Xue-yong Fu, Mahsa Azizi, Shashi Bhushan, and Simon Corston-oliver. 2023{\natexlab{b}}.
\newblock \href {https://doi.org/10.18653/v1/2023.acl-industry.57} {{AI} coach assist: An automated approach for call recommendation in contact centers for agent coaching}.
\newblock In \emph{Proceedings of the 61st Annual Meeting of the Association for Computational Linguistics (Volume 5: Industry Track)}, pages 599--607, Toronto, Canada. Association for Computational Linguistics.

\bibitem[{Laskar et~al.(2022{\natexlab{a}})Laskar, Chen, Johnston, Fu, Bhushan~TN, and Corston-Oliver}]{laskar2022auto}
Md~Tahmid~Rahman Laskar, Cheng Chen, Jonathan Johnston, Xue-Yong Fu, Shashi Bhushan~TN, and Simon Corston-Oliver. 2022{\natexlab{a}}.
\newblock An auto encoder-based dimensionality reduction technique for efficient entity linking in business phone conversations.
\newblock In \emph{Proceedings of the 45th International ACM SIGIR Conference on Research and Development in Information Retrieval}, pages 3363--3367.

\bibitem[{Laskar et~al.(2022{\natexlab{b}})Laskar, Chen, Martsinovich, Johnston, Fu, Tn, and Corston-Oliver}]{laskar-etal-2022-blink}
Md~Tahmid~Rahman Laskar, Cheng Chen, Aliaksandr Martsinovich, Jonathan Johnston, Xue-Yong Fu, Shashi~Bhushan Tn, and Simon Corston-Oliver. 2022{\natexlab{b}}.
\newblock \href {https://doi.org/10.18653/v1/2022.naacl-industry.38} {{BLINK} with {E}lasticsearch for efficient entity linking in business conversations}.
\newblock In \emph{Proceedings of the 2022 Conference of the North American Chapter of the Association for Computational Linguistics: Human Language Technologies: Industry Track}, pages 344--352, Hybrid: Seattle, Washington + Online. Association for Computational Linguistics.

\bibitem[{Laskar et~al.(2022{\natexlab{c}})Laskar, Hoque, and Huang}]{laskar2022domain}
Md~Tahmid~Rahman Laskar, Enamul Hoque, and Jimmy~Xiangji Huang. 2022{\natexlab{c}}.
\newblock Domain adaptation with pre-trained transformers for query-focused abstractive text summarization.
\newblock \emph{Computational Linguistics}, 48(2):279--320.

\bibitem[{Lewis et~al.(2020)Lewis, Liu, Goyal, Ghazvininejad, Mohamed, Levy, Stoyanov, and Zettlemoyer}]{bart}
Mike Lewis, Yinhan Liu, Naman Goyal, Marjan Ghazvininejad, Abdelrahman Mohamed, Omer Levy, Veselin Stoyanov, and Luke Zettlemoyer. 2020.
\newblock \href {https://doi.org/10.18653/v1/2020.acl-main.703} {{BART}: Denoising sequence-to-sequence pre-training for natural language generation, translation, and comprehension}.
\newblock In \emph{Proceedings of the 58th Annual Meeting of the Association for Computational Linguistics}, pages 7871--7880, Online. Association for Computational Linguistics.

\bibitem[{Lin(2004)}]{lin2004rouge}
Chin-Yew Lin. 2004.
\newblock Rouge: A package for automatic evaluation of summaries.
\newblock In \emph{Text summarization branches out}, pages 74--81.

\bibitem[{Liu et~al.(2023)Liu, Lin, Hewitt, Paranjape, Bevilacqua, Petroni, and Liang}]{liu2023lost}
Nelson~F Liu, Kevin Lin, John Hewitt, Ashwin Paranjape, Michele Bevilacqua, Fabio Petroni, and Percy Liang. 2023.
\newblock Lost in the middle: How language models use long contexts.
\newblock \emph{arXiv preprint arXiv:2307.03172}.

\bibitem[{Manderscheid and Lee(2023)}]{manderscheid-lee-2023-predicting}
Etienne Manderscheid and Matthias Lee. 2023.
\newblock \href {https://doi.org/10.18653/v1/2023.acl-industry.62} {Predicting customer satisfaction with soft labels for ordinal classification}.
\newblock In \emph{Proceedings of the 61st Annual Meeting of the Association for Computational Linguistics (Volume 5: Industry Track)}, pages 652--659, Toronto, Canada. Association for Computational Linguistics.

\bibitem[{Nedoluzhko et~al.(2022)Nedoluzhko, Singh, Hled{\'\i}kov{\'a}, Ghosal, and Bojar}]{nedoluzhko2022elitr}
Anna Nedoluzhko, Muskaan Singh, Marie Hled{\'\i}kov{\'a}, Tirthankar Ghosal, and Ond{\v{r}}ej Bojar. 2022.
\newblock Elitr minuting corpus: A novel dataset for automatic minuting from multi-party meetings in english and czech.
\newblock In \emph{Proceedings of the Thirteenth Language Resources and Evaluation Conference}, pages 3174--3182.

\bibitem[{OpenAI(2023)}]{openai2023gpt4}
OpenAI. 2023.
\newblock \href {http://arxiv.org/abs/2303.08774} {Gpt-4 technical report}.

\bibitem[{Qin et~al.(2023)Qin, Zhang, Zhang, Chen, Yasunaga, and Yang}]{qin2023chatgpt}
Chengwei Qin, Aston Zhang, Zhuosheng Zhang, Jiaao Chen, Michihiro Yasunaga, and Diyi Yang. 2023.
\newblock Is chatgpt a general-purpose natural language processing task solver?
\newblock \emph{arXiv preprint arXiv:2302.06476}.

\bibitem[{Raffel et~al.(2020)Raffel, Shazeer, Roberts, Lee, Narang, Matena, Zhou, Li, and Liu}]{t5}
Colin Raffel, Noam Shazeer, Adam Roberts, Katherine Lee, Sharan Narang, Michael Matena, Yanqi Zhou, Wei Li, and Peter~J Liu. 2020.
\newblock Exploring the limits of transfer learning with a unified text-to-text transformer.
\newblock \emph{The Journal of Machine Learning Research}, 21(1):5485--5551.

\bibitem[{Touvron et~al.(2023)Touvron, Martin, Stone, Albert, Almahairi, Babaei, Bashlykov, Batra, Bhargava, Bhosale et~al.}]{touvron2023llama2}
Hugo Touvron, Louis Martin, Kevin Stone, Peter Albert, Amjad Almahairi, Yasmine Babaei, Nikolay Bashlykov, Soumya Batra, Prajjwal Bhargava, Shruti Bhosale, et~al. 2023.
\newblock Llama 2: Open foundation and fine-tuned chat models.
\newblock \emph{arXiv preprint arXiv:2307.09288}.

\bibitem[{Vaswani et~al.(2017)Vaswani, Shazeer, Parmar, Uszkoreit, Jones, Gomez, Kaiser, and Polosukhin}]{DBLP:conf/nips/VaswaniSPUJGKP17}
Ashish Vaswani, Noam Shazeer, Niki Parmar, Jakob Uszkoreit, Llion Jones, Aidan~N. Gomez, Lukasz Kaiser, and Illia Polosukhin. 2017.
\newblock \href {https://proceedings.neurips.cc/paper/2017/hash/3f5ee243547dee91fbd053c1c4a845aa-Abstract.html} {Attention is all you need}.
\newblock In \emph{Advances in Neural Information Processing Systems 30: Annual Conference on Neural Information Processing Systems 2017, December 4-9, 2017, Long Beach, CA, {USA}}, pages 5998--6008.

\bibitem[{Wolf et~al.(2020)Wolf, Debut, Sanh, Chaumond, Delangue, Moi, Cistac, Rault, Louf, Funtowicz et~al.}]{wolf2019huggingface}
Thomas Wolf, Lysandre Debut, Victor Sanh, Julien Chaumond, Clement Delangue, Anthony Moi, Pierric Cistac, Tim Rault, R{\'e}mi Louf, Morgan Funtowicz, et~al. 2020.
\newblock Transformers: State-of-the-art natural language processing.
\newblock In \emph{Proceedings of the 2020 conference on empirical methods in natural language processing: system demonstrations}, pages 38--45.

\bibitem[{Zhang et~al.(2020)Zhang, Zhao, Saleh, and Liu}]{pegasus}
Jingqing Zhang, Yao Zhao, Mohammad Saleh, and Peter Liu. 2020.
\newblock Pegasus: Pre-training with extracted gap-sentences for abstractive summarization.
\newblock In \emph{International Conference on Machine Learning}, pages 11328--11339. PMLR.

\bibitem[{Zhang et~al.(2023)Zhang, Dong, Li, Zhang, Sun, Wang, Li, Hu, Zhang, Wu et~al.}]{zhang2023instruction}
Shengyu Zhang, Linfeng Dong, Xiaoya Li, Sen Zhang, Xiaofei Sun, Shuhe Wang, Jiwei Li, Runyi Hu, Tianwei Zhang, Fei Wu, et~al. 2023.
\newblock Instruction tuning for large language models: A survey.
\newblock \emph{arXiv preprint arXiv:2308.10792}.

\bibitem[{Zhang et~al.(2019)Zhang, Kishore, Wu, Weinberger, and Artzi}]{zhang2019bertscore}
Tianyi Zhang, Varsha Kishore, Felix Wu, Kilian~Q Weinberger, and Yoav Artzi. 2019.
\newblock Bertscore: Evaluating text generation with bert.
\newblock In \emph{International Conference on Learning Representations}.

\bibitem[{Zhao et~al.(2023)Zhao, Zhou, Li, Tang, Wang, Hou, Min, Zhang, Zhang, Dong et~al.}]{zhao2023survey}
Wayne~Xin Zhao, Kun Zhou, Junyi Li, Tianyi Tang, Xiaolei Wang, Yupeng Hou, Yingqian Min, Beichen Zhang, Junjie Zhang, Zican Dong, et~al. 2023.
\newblock A survey of large language models.
\newblock \emph{arXiv preprint arXiv:2303.18223}.

\bibitem[{Zhong et~al.(2021)Zhong, Yin, Yu, Zaidi, Mutuma, Jha, Hassan, Celikyilmaz, Liu, Qiu et~al.}]{zhong2021qmsum}
Ming Zhong, Da~Yin, Tao Yu, Ahmad Zaidi, Mutethia Mutuma, Rahul Jha, Ahmed Hassan, Asli Celikyilmaz, Yang Liu, Xipeng Qiu, et~al. 2021.
\newblock Qmsum: A new benchmark for query-based multi-domain meeting summarization.
\newblock In \emph{Proceedings of the 2021 Conference of the North American Chapter of the Association for Computational Linguistics: Human Language Technologies}, pages 5905--5921.

\bibitem[{Zhu et~al.(2021)Zhu, Liu, Mei, and Zeng}]{zhu2021mediasum}
Chenguang Zhu, Yang Liu, Jie Mei, and Michael Zeng. 2021.
\newblock \href {https://doi.org/10.18653/v1/2021.naacl-main.474} {{M}edia{S}um: A large-scale media interview dataset for dialogue summarization}.
\newblock In \emph{Proceedings of the 2021 Conference of the North American Chapter of the Association for Computational Linguistics: Human Language Technologies}, pages 5927--5934, Online. Association for Computational Linguistics.

\bibitem[{Zhu et~al.(2023)Zhu, Li, Liu, Ma, and Wang}]{zhu2023survey}
Xunyu Zhu, Jian Li, Yong Liu, Can Ma, and Weiping Wang. 2023.
\newblock A survey on model compression for large language models.
\newblock \emph{arXiv preprint arXiv:2308.07633}.

\end{thebibliography}
\bibliographystyle{acl_natbib}

\end{document}